\documentclass{article}
\usepackage{spconf,amsmath,graphicx}
\usepackage{enumitem}
\usepackage{booktabs}  
\usepackage{siunitx}
\usepackage{multirow}
\usepackage{adjustbox} 
\usepackage{pifont}
\usepackage[hidelinks]{hyperref}   
\usepackage[caption=false,font=footnotesize]{subfig}

\usepackage[table]{xcolor}
\newcommand{\cmark}{\ding{51}}

\usepackage{setspace}
\usepackage{etoolbox}
\apptocmd{\thebibliography}{%
  \setlength{\itemsep}{0pt}
  \setlength{\parskip}{0pt}
}{}{}

\title{LOTUSDIS: A Thai Far-Field Meeting Corpus for Robust CONVERSATIONAL~ASR}

\name{Pattara Tipaksorn, 
        Sumonmas Thatphithakkul,
        Vataya Chunwijitra, 
        Kwanchiva Thangthai}
\address{Speech and Text Understanding Research Team, NECTEC \\
        \{pattara.tip, sumonmas.tha, vataya.chu, kwanchiva.tha\}@nectec.or.th}

\begin{document}
%

\maketitle
%


\begin{abstract}
We present LOTUSDIS, a publicly available Thai meeting corpus designed to advance far-field conversational ASR. The dataset comprises 114 hours of spontaneous, unscripted dialogue collected in 15–20 minute sessions with three participants, where overlapping speech is frequent and natural. Speech was recorded simultaneously by nine independent single-channel devices spanning six microphone types at distances from ~0.12 m to 10 m, preserving the authentic effects of reverberation, noise, and device coloration without relying on microphone arrays. We provide standard train/dev/test splits and release a reproducible baseline system. We benchmarked several Whisper variants under zero-shot and fine-tuned conditions. Off-the-shelf models showed strong degradation with distance, confirming a mismatch between pre-training data and Thai far-field speech. Fine-tuning on LOTUSDIS dramatically improved robustness: a Thai Whisper baseline reduced overall WER from 64.3\% to 38.3\% and far-field WER from 81.6\% to 49.5\%, with especially large gains on the most distant microphones. These results underscore the importance of distance-diverse training data for robust ASR. The corpus is available under CC-BY-SA 4.0\textsuperscript{1}\footnotetext[1]{https://github.com/kwanchiva/LOTUSDIS}. We also release a training and evaluation scripts as a baseline system to promote reproducible research in this field.

\end{abstract}
\section{Introduction}

Automatic speech recognition (ASR) has advanced rapidly, yet performance remains fragile in real-world meeting scenarios. Applications such as meeting transcription demand robustness to overlapping speech and diverse microphone setups, conditions that are still difficult for current systems. While distant speech recognition (DSR) has been studied extensively in English and other high-resource languages, Thai resources remain limited. Most available corpora focus on near-field read speech~\cite{suwanbandit2023thai,ardila-etal-2020-common,lotus-soc,lotusted,conneau2023fleurs} or broadcast news~\cite{lotus-bn}, making them unsuitable for modeling far-field conversational dynamics. In addition, restrictive licensing practices~\cite{lotus-soc,lotusted,lotus-bn} further hinder reproducibility and fair benchmarking.

Here, we introduce LOTUSDIS, the first publicly available Thai far-field conversational speech corpus. The dataset captures natural, unscripted meetings with multiple microphone types and distances, offering a realistic testbed without relying on costly microphone arrays. Figure 1 illustrates the overall meeting setup in LOTUSDIS along with example spectrograms from each microphone. LOTUSDIS is designed to support both model development and fair benchmarking for Thai distant speech recognition. This work makes three main contributions:

\noindent\textbf{Open corpus}: We release LOTUSDIS, a 114-hour Thai meeting corpus under a permissive CC-BY-SA 4.0 license.

\noindent\textbf{Reproducible baselines}: We provide a reproducible benchmark by fine-tuning Whisper on LOTUSDIS, establishing a strong baseline for future Thai DSR research.

\noindent\textbf{Empirical insights}: We present a systematic analysis of ASR challenges, including the impact of microphone distance, training with single-microphone data, and the effect of naturally occurring speaker overlap.

\begin{figure}[!t]
  \centering
  \subfloat[Room layout with speaker/microphone positions.\label{fig:layout}]{
    \includegraphics[width=\linewidth]{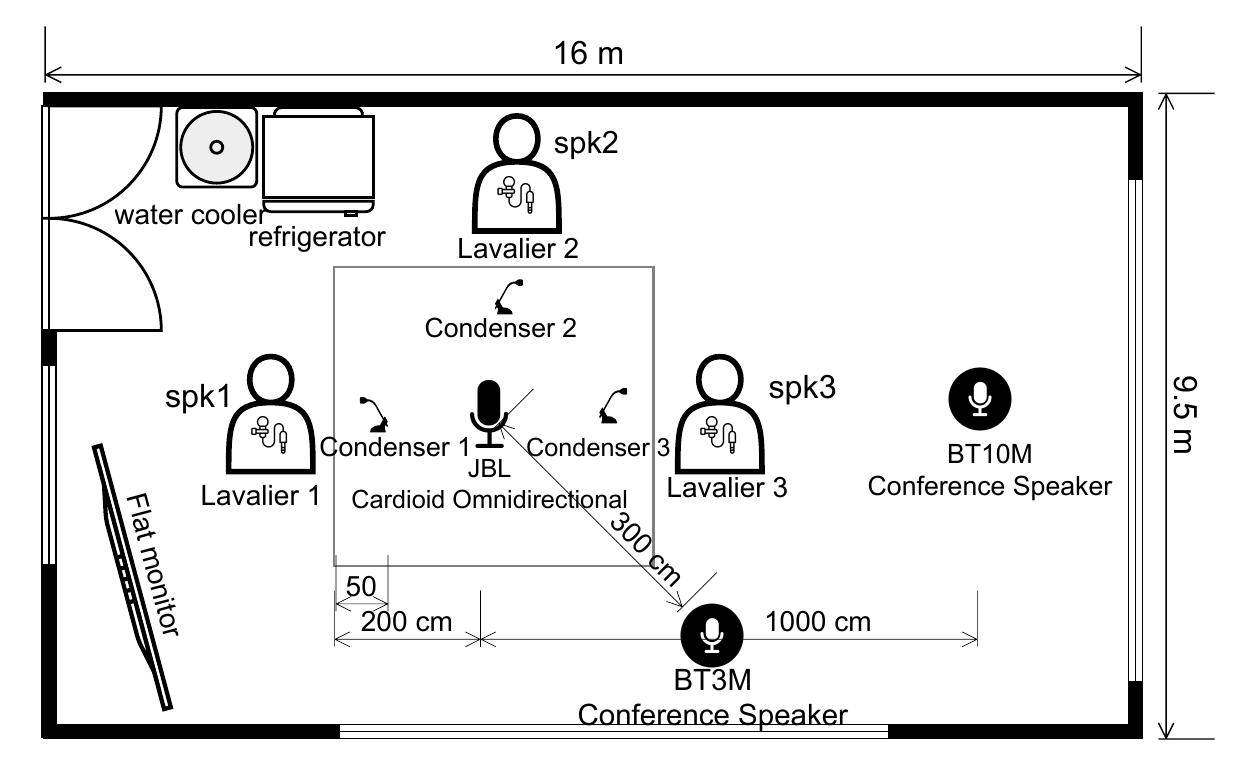}
  }\vspace{0.3em}
  \subfloat[Example spectrograms across microphone types.\label{fig:spectrograms}]{
    \includegraphics[width=\linewidth]{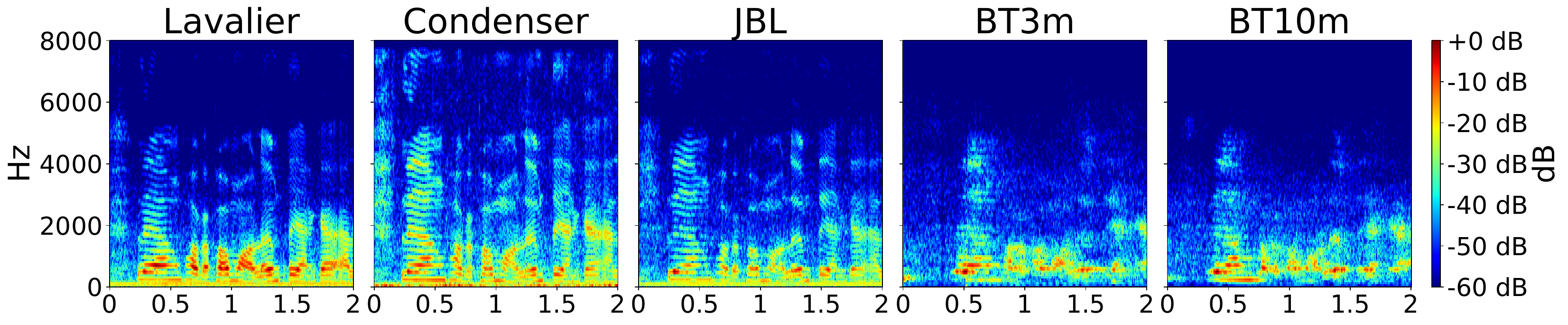}
  }
  \caption{Overview of the LOTUSDIS dataset setup and microphone characteristics.}
  \label{fig:lotusdis_overview}
\end{figure}

\begin{table*}[t]
  \centering
  \caption{Comparison of meeting-style / far-field corpora. Distance coverage is approximate talker–mic range.}
  \label{tab:corpora_compare}
  \begin{adjustbox}{width=\linewidth}
  \begin{tabular}{l l l c l l l l l}
    \toprule
    \textbf{Dataset} & \textbf{Language} & \textbf{Hours} & \textbf{Far-field} & \textbf{\#Sessions} & \textbf{Speakers/session} & \textbf{\#Speakers} & \textbf{Distance coverage} & 
    \textbf{License}\\
    \midrule
    AMI Meeting Corpus~\cite{carletta2007unleashing} & English & 100 & \cmark & $\approx$137 & 4 &  -- & 0.3--3 m close-talk + mic arrays  & CC BY 4.0 \\
    ICSI Meeting Corpus~\cite{1198793} & English & 72 & \cmark & 75 & 3--10 &  -- & close-talk worn + table top & CC BY 4.0\\
    AliMeeting~\cite{Yu2022M2MeT} & Chinese & 120 & \cmark & 220 & 2--4 &  481 & 0.3--5 m headset + circular array & CC BY-SA 4.0\\
    LibriCSS~\cite{libricss} & English & 10 & \cmark & 10 & 8 &  40 & 0.3–4 m circular array (audio play back)   & CC BY 4.0 \\
    CHiME-6 (CHiME-5 data)~\cite{barker18_interspeech,watanabe2020chime} & English & 50 & \cmark & 20 & 4 &  48 & 0.2–4 m close-talk + linear arrays & CC BY-SA 4.0 \\
    NOTSOFAR-1~\cite{vinnikov24_interspeech} & English & 28 (260 all mics) & \cmark & 315 & 4-8 &  35 & close talk + far-field & CC BY-NC-ND 4.0\\
    DiPCo~\cite{segbroeck20_interspeech} & English & 5 & \cmark & 10 & 4 &  32 & 1–4 m close-talk  + circular arrays & CDLA-Permissive\\
     AISHELL-4~\cite{fu21b_interspeech} & Chinese & 120 & \cmark & 211 & 4--8 &  61 & 0.6--6 m headset + circular array & CC BY-SA 4.0\\
    \rowcolor{black!6}
    \textbf{LOTUSDIS (ours)} & \textbf{Thai} & \textbf{20 (114 all mics)} & \textbf{\cmark} & \textbf{90} & \textbf{3} &  \textbf{86} & \textbf{0.12--10 m lavalier + table top} & CC BY-SA 4.0\\
    \bottomrule
  \end{tabular}
  \end{adjustbox}
\end{table*}

\section{Related Work and Corpus Comparison}
The development of speech corpora has been central to advances in ASR, with several large-scale resources released for meeting and conversational scenarios, especially in English and Chinese. These datasets have enabled progress in speech separation, diarization, and distant speech recognition. The AMI Meeting Corpus~\cite{carletta2007unleashing} and the ICSI Meeting Corpus~\cite{1198793} are among the most influential English resources, capturing multi-party interactions in realistic meeting rooms. However, despite their multi-microphone setups, many recordings rely heavily on close-talk or headset microphones, which reduces their applicability to true far-field recognition tasks.

More recent initiatives have specifically targeted the far-field challenge. The CHiME-6~\cite{barker18_interspeech} and AISHELL-4~\cite{fu21b_interspeech} challenges introduced datasets with multi-channel array recordings in realistic domestic and conference settings. Similarly, the AliMeeting corpus~\cite{Yu2022M2MeT} and DiPCo~\cite{segbroeck20_interspeech} provide large-scale Chinese and English meeting data with headset and circular array recordings. The NOTSOFAR-1 Challenge~\cite{vinnikov24_interspeech} is another English corpus with 280 sessions featuring 4–8 speakers each, offering both close-talk and far-field recordings. These corpora have been highly effective in fostering research on beamforming, source separation, and array processing. However, their reliance on microphone arrays presents a different set of research problems and may not reflect more common deployment scenarios where only single-channel, off-the-shelf devices are available.

LOTUSDIS addresses this gap by offering a non-array, single-channel evaluation framework across diverse microphone types and distances. Unlike array-focused corpora such as AliMeeting and AISHELL-4, LOTUSDIS targets a low-resource language and emphasizes realistic, deployment-oriented settings. This design enables research on robust ASR under practical far-field conditions without the need for specialized hardware. A comparative summary of LOTUSDIS and other prominent meeting-style corpora is provided in Table~\ref{tab:corpora_compare}, highlighting its unique position in the research landscape.

\section{The LOTUSDIS Corpus}

\subsection{Data Collection Methodology}

The LOTUSDIS corpus was designed to capture realistic conversational speech in an authentic office environment. Recordings were conducted in a furnished office room measuring 16 × 9.5 × 2.7 m (length × width × height). To preserve ecological validity, the Heating Ventilation and Air Conditioning (HVAC) system, along with common appliances such as a water cooler and refrigerator, operated normally during all sessions. This intentional inclusion of stationary background noise creates a natural noise floor representative of real-world meeting conditions, which is crucial for ASR.

\begin{table}[!t]
  \centering
  \caption{Details of LOTUSDIS dataset.}
  \label{tab:lotusdis_details}
  \begin{adjustbox} {width=0.65\linewidth}
  
  \begin{tabular}{lrrr}
    \toprule
    & \textbf{Train} & \textbf{Dev} & \textbf{Test} \\
    \midrule
    Duration (hh:mm)          & 17:37 & 2:33 & 2:39 \\
    5-mic total duration          & 88:07 & 12:49 & 13:17 \\
    \#Session             & 69    & 10    & 11    \\
    \#Topic               & 40    & 10    & 11    \\
    \#Participant         & 74    & 12    & 12    \\
    \#Male                & 35    & 5     & 7     \\
    \#Female              & 39    & 7     & 5     \\
    \#Utterances          & 120{,}245 & 13{,}090 & 27{,}580 \\
    Overlap Ratio (Avg.)  & 31.5\% & 40.3\% & 29.0\% \\
    \bottomrule
  \end{tabular}
\end{adjustbox}
\end{table}

In total, the corpus contains 114 hours of multi-channel speech, derived from approximately 20 hours of unique meeting sessions. The data is partitioned into 88 hours for training, 12.8 hours for development, and 13.3 hours for evaluation. Each session lasted about 15 minutes and involved three participants engaged in spontaneous, unscripted conversation prompted by a pre-assigned topic. The free-flowing nature of the discussions naturally produced frequent overlap, a defining feature of the dataset. Across all recordings, 86 unique participants were involved, ranging in age from 19 to 48 years, with an average age of 27. As summarized in Table~\ref{tab:lotusdis_details}, roughly one-third of the total audio consists of overlapping speech, underscoring the dataset’s importance for research on challenging multi-speaker scenarios.

\begin{table*}[!t]
\centering
\caption{WER (\%) by microphone and training recipe. Near-/Far-field averages and Overall include all listed mics.}
\label{tab:main_results}
\small
\setlength{\tabcolsep}{4pt}
\begin{adjustbox}{width=\linewidth}
\begin{tabular}{@{}lllccrrrrr@{}} 
\toprule
\multicolumn{3}{c}{} &
\multicolumn{2}{c}{\textbf{Near-field ($\le$ 1 m)}} &
\multicolumn{3}{c}{\textbf{Far-field (2--10 m)}} &
\multicolumn{1}{c}{\textbf{Macro Avg}} \\
\cmidrule(lr){4-5}\cmidrule(lr){6-8}\cmidrule(lr){9-10}
\textbf{Base model} & \textbf{Fine-tuned data} & \textbf{Front-end} &
\textbf{lavalier} & \textbf{condenser} &
\textbf{jbl} & \textbf{bt3m} & \textbf{bt10m} &
\textbf{Near-/Far-field} & \textbf{Overall} \\
\midrule
\multicolumn{10}{@{}l}{\emph{(A) Zero-shot}}\\
Whisper-large-v3~\cite{radford2022whisper} & --      & --   & {51.95} & {48.37} & {59.90} & {117.03} & {125.52} & {50.16/100.82} & {79.84} \\
Pathumma-whisper-th-large-v3~\cite{tipaksorn2024PathummaWhisper,10800399} & --      & --   & {37.55} & {36.43} & {44.22} & {96.27} & {104.22} & {36.99/81.57~~} & {64.32} \\
Biodatlab/whisper-th-large-v3-combined~\cite{aung2024thonburian} & --      & --   & {42.51} & {39.45} & {46.00} & {97.77} & {106.43} & {40.98/83.40~~} & {66.36} \\
Monsoon-whisper-medium-gigaspeech2~\cite{monsoon} & --      & --   & {39.03} & {37.29} & {43.31} & {109.10} & {124.87} & {38.16/92.43~~} & {66.15} \\
\midrule
\addlinespace[1mm]
\multicolumn{10}{@{}l}{\emph{(B) Full Fine-tuned on All Mic (The Baseline)}}\\
Whisper-large-v3  & All Mic & --   &   23.20 & 20.81 & 27.27 & 59.00 & 65.25 & 22.01/50.51~~ & 39.05 \\
Pathumma-whisper-th-large-v3   & All Mic & --   & 22.77 & 20.40 & \bfseries 26.42 &  \bfseries 58.15 & \bfseries 64.04 &   21.59/\bfseries49.54~~ &  \bfseries 38.33 \\

\midrule
\addlinespace[1mm]
\multicolumn{10}{@{}l}{\emph{(C) Front-end processing}}\\
Pathumma-whisper-th-large-v3   & All Mic & WPE \cite{Drude2018NaraWPE}  & 37.14 & 34.69 & 37.00 & 63.04 & 68.32 & 35.92/56.12~~ & 48.00 \\
Pathumma-whisper-th-large-v3  & All Mic & MMSE-LSA~\cite{ephraim2003speech} & {26.69} & {23.14} & {31.22} & {62.92} & {69.52} & {24.92/54.55~~} & {42.89} \\
\midrule
\addlinespace[1mm]

\multicolumn{10}{@{}l}{\emph{(D) Single-mic fine-tunes}}\\
Pathumma-whisper-th-large-v3  & Condenser & --  & 22.27 & 19.26 & 27.02 & 97.95 & 113.65 & {20.77}/79.54~~ & 50.12 \\
Pathumma-whisper-th-large-v3   & BT3M     & --  & 29.38 & 26.61 & 34.51 & 60.17 & 69.86 & 28.00/54.85~~ & 44.75 \\ 
\midrule
\addlinespace[1mm]
\multicolumn{10}{@{}l}{\emph{(E) Data augmentation (single-mic)}}\\
Pathumma-whisper-th-large-v3   & Condenser+Spec~\cite{park19e_interspeech} & -- & {23.16} & {19.67} & {28.99} & {82.40} & {90.06} & {21.42/67.15~~} & {49.11} \\

Pathumma-whisper-th-large-v3  & Condenser+Reverb & --& \textbf{21.57} & \textbf{18.77} & {26.61} & {80.30} & {89.25} & \textbf{20.17}/65.39~~ & {45.86} \\

\bottomrule
\end{tabular}
\end{adjustbox}
\end{table*}
\vspace{-2mm}

\subsection{Microphone Setup and Spatial Configuration}
A distinguishing feature of LOTUSDIS is its simultaneous multi-channel recording setup without reliance on microphone arrays. Instead, nine independent single-channel microphones representing six device types were deployed to capture speech across a wide range of acoustic conditions. The spatial configuration of the recording sessions is shown in Fig~\ref{fig:lotusdis_overview}\subref{fig:layout}. Microphones were grouped into near-field and far-field zones, providing paired reference and degraded signals that expose ASR systems to a full spectrum of acoustic paths and device colorations. As illustrated in the spectrograms in Fig~\ref{fig:lotusdis_overview}\subref{fig:spectrograms}, signal quality visibly deteriorates with increasing distance: near-field channels retain clear high-frequency formants, while far-field devices such as BT10m exhibit pronounced high-frequency roll-off, lower SNR, and stronger reverberation.

\noindent\textbf{Near-field ($\leq$0.5 m).}
Three lavaliers and three table-mounted condensers, positioned 12–15 cm from each speaker’s mouth, provide high-quality, high-DRR (direct-to-reverberant ratio) reference recordings.

\noindent\textbf{Far-field ($\geq$2 m).}
A tabletop JBL loudspeaker (2 m) and two Bluetooth speakerphones placed at 3 m (BT3m) and 10 m (BT10m) capture increasingly reverberant, low-SNR speech with marked spectral tilt.

All microphones were fixed at measured distances with line-of-sight maintained. Synchronization across channels was achieved with slate pulses and verified by cross-correlation, ensuring sub-sample alignment and minimal drift throughout each session. This diverse, multi-device configuration offers a rich testbed for assessing ASR robustness to distance, reverberation, and device coloration without array processing.





\vspace{-0.4em}
\subsection{Annotation and Metadata}
\vspace{-0.4em}
The annotation of LOTUSDIS followed a two-stage protocol to ensure consistency and benchmark-quality labels. In the first stage, three trained annotators segmented and transcribed the recordings according to a unified guideline covering tokenization, code-switching, and the treatment of non-lexical events. In the second stage, a senior annotator reviewed every session to resolve discrepancies, correct boundary errors, and validate the transcripts. The final corpus provides utterance-level transcripts with speaker labels and an explicit overlap mask, enabling controlled evaluation of ASR systems under multi-speaker conditions. A lightweight Thai tag set was adopted, including $<$n$>$ for noise, $<$sil$>$ for silences exceeding 300 ms, $<$unk$>$ for unintelligible spans, and $<$td$>$ for dialectal uncertainty. Speaker overlaps are indicated by concatenating speaker IDs with an ampersand.



\section{Experiments and Empirical Analysis}
\subsection{Experimental Setup}
To assess the utility of LOTUSDIS, we conducted a set of baseline ASR experiments using multiple variants of the Whisper architecture. As the primary reference system, we adopted Pathumma-whisper-th-large-v3~\cite{tipaksorn2024PathummaWhisper}, a Thai-specific fine-tuned model. For comparability, all conditions shared the same decoding, text normalization, and scoring pipeline, with word tokenization performed using the newmm segmenter from PyThaiNLP~\cite{pythainlp}. Performance is reported in terms of Word Error Rate (WER), with results broken down by microphone type. In addition to per-microphone scores, we report macro-averages for near-field versus far-field conditions, as well as a global average across all channels. Fine-tuning was performed on the LOTUSDIS training split for five epochs using a single NVIDIA H200 GPU.

\subsection{Zero-Shot Performance and Domain Mismatch}

The initial experiments evaluated off-the-shelf, zero-shot Whisper models on the LOTUSDIS test set without any domain adaptation. The results are shown in Table~\ref{tab:main_results}(a). As anticipated, performance degraded markedly with increasing microphone distance, reflecting a strong mismatch between the models’ pre-training distribution and the acoustic properties of Thai far-field recordings. Across all microphones, WER increased systematically from near-field lavaliers and condensers to the far-field JBL, BT3m, and BT10m devices. These results confirm that the zero-shot performance of large pre-trained models is highly sensitive to the acoustic mismatch between training and target conditions.

\subsection{Fine-Tuned Baselines} \vspace{-0.4em}

The second experiments revealed the substantial benefits of fine-tuning on LOTUSDIS, shown in Table~\ref{tab:main_results}(b). For Whisper-large-v3, fine-tuning reduced the overall WER to 39.05\% and the far-field average to 50.51\%, with microphone-specific improvements such as a drop to 59.00\% on BT3m and 65.25\% on BT10m. Pathumma-whisper-th-large-v3 showed even stronger gains, with the overall WER reduced from 64.3\% in the zero-shot condition to 38.33\% after fine-tuning. Far-field performance improved from 81.6\% to 49.54\%, while the BT3m and BT10m microphones dropped to 58.15\% and 64.04\%, respectively. These findings underscore that exposure to distance-diverse training data not only improves overall accuracy but also mitigates the steep degradation otherwise observed under far-field conditions, thereby validating LOTUSDIS as an effective resource for building robust ASR systems in real-world acoustic environments.

\vspace{-0.4em}
\subsection{Speech Front-end Processing} \vspace{-0.4em}

We further examined the impact of traditional single-channel front-end processing on recognition performance. As shown in Table~\ref{tab:main_results}(C), two methods were evaluated: Weighted Prediction Error (WPE) dereverberation, implemented with the NaraWPE Python package~\cite{Drude2018NaraWPE}, and Minimum Mean-Square Error Log-Spectral Amplitude (MMSE-LSA) spectral subtraction~\cite{ephraim2003speech}. The results indicate that such processing is not universally beneficial. Applying WPE consistently degraded performance, increasing near-field WER from 21.59\% to 35.92\% and far-field WER from 49.54\% to 56.12\%. Although WPE is effective for suppressing late reverberation, it can also introduce artifacts that distort non-reverberant signals. MMSE-LSA showed a similar trend, raising near-field WER to 24.92\% and far-field WER to 54.55\%. These findings highlight an important practical insight: a uniform front-end strategy is suboptimal. In this setting, fine-tuned models trained directly on distance-diverse data achieved greater robustness than either dereverberation or denoising. More effective deployment may require applying front-end processing selectively to far-field channels or adopting adaptive strategies that account for input acoustic characteristics.

\vspace{-0.4em}
\subsection{Single-Mic Fine-Tuning and Augmentation}  \vspace{-0.4em}
In many real-world applications, data collection may be restricted to a single microphone type. To examine this constraint, we fine-tuned models using only one channel, as summarized in Table~\ref{tab:main_results}(D). A model trained exclusively on the near-field condenser microphone achieved a WER of 19.26\% on that device—surpassing the full all-microphone baseline in-domain. However, this came at the cost of extreme degradation on unseen far-field channels, with BT3m performance collapsing to 97.95\%. These results confirm that single-channel training leads to severe overfitting to device-specific acoustics, undermining generalization to new conditions. To mitigate this limitation, we explored lightweight, physically motivated data augmentation, as shown in Table~\ref{tab:main_results}(E). Augmentations included convolution with real room impulse responses (RIRs) from OpenSLR-28~\cite{RIR} and additive mixing with HVAC and conversational noise, simulating far-field degradations. This approach substantially improved robustness: for the condenser-trained model, reverberation-based augmentation reduced far-field WER from 79.54\% to 65.39\%, yielding the best performance among single-microphone systems. These findings demonstrate that simple augmentation can effectively bridge the gap when training data is limited to one microphone type, preventing overfitting and improving generalization.

\vspace{-0.4em}
\subsection{The Challenge of Overlapped Speech} 
\begin{figure}[!tbp]
        \centering
        \includegraphics[width=0.8\linewidth]{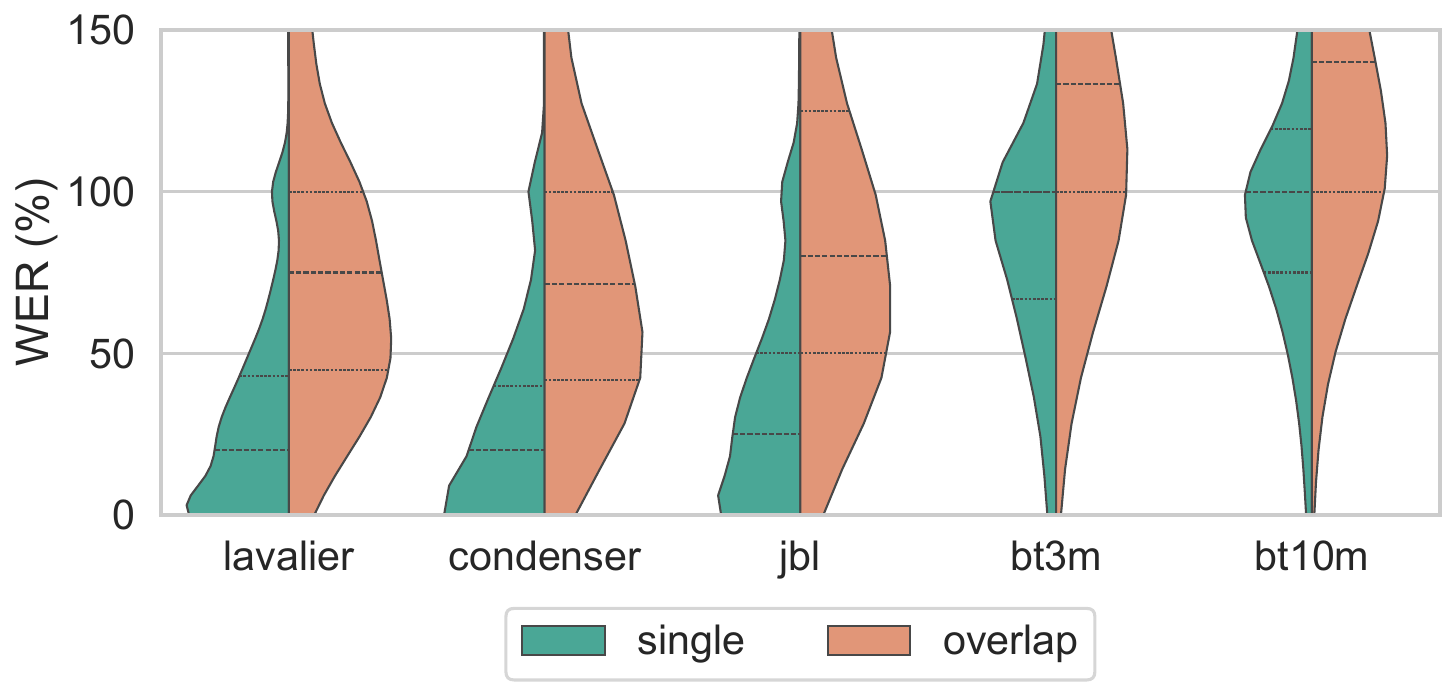}
        \vspace{-0.4em} \caption{Utterance-level WER distributions for single-speaker vs overlapped speech across microphones (y-axis clipped at 150\%). Each half-violin shows the test-set distribution and dashed lines mark quartiles. }
        \label{fig:overlapspk}
    \vspace{-0.4em}
    \end{figure}

\vspace{-0.4em}
We conducted a focused analysis of overlapped speech, a pervasive challenge in conversational data. As shown in Fig~\ref{fig:overlapspk}, the WER distribution shifts upward for all microphones under overlap conditions, with the effect most severe on long-range devices such as BT3m and BT10m. This reflects a strong overlap–distance interaction, where reduced signal quality compounds the difficulty of multi-speaker recognition. Detailed error analysis revealed an increase in deletions and short-token drops, suggesting that the system often fails to track the target speaker. Additionally, substitution errors were more frequent in Thai’s tone-bearing syllables, indicating sensitivity to tonal cues under masking. These findings highlight overlapping speech as a critical obstacle for far-field ASR, underscoring the need for advances in both speech separation and linguistic modeling.

\vspace{-0.4em}
\section{Conclusion} \vspace{-0.4em}

This paper introduced LOTUSDIS, a Thai meeting corpus designed to advance distant speech recognition without relying on beamforming arrays. As the first publicly available, open-access dataset for a low-resource language, LOTUSDIS fills a critical gap by providing realistic, unscripted conversation across a wide range of single-channel microphones and distances. Our empirical analysis serves as a foundational benchmark, demonstrating the corpus's value: fine-tuning a Thai Whisper model on LOTUSDIS resulted in a substantial reduction of the Overall WER from 64.3\% to 38.3\%, validating it as an essential resource for robust DSR systems. The work also provides practical insights into addressing data scarcity through augmentation and highlights the persistent challenge of speaker overlap.

\vfill\pagebreak
\bibliographystyle{IEEEtran} 
\bibliography{refs}


\end{document}